\documentclass[twoside]{article}

\usepackage[accepted]{aistats2020}
\usepackage[round]{natbib}

\usepackage[utf8]{inputenc} 
\usepackage[T1]{fontenc}    
\usepackage[parfill]{parskip}
\usepackage
{  hyperref,
	url,
	booktabs,
	amsfonts,
	nicefrac,
	microtype,
	bbm,
	graphicx,
	amssymb,
	amsmath,
	amsthm,
	dsfont, 
	xcolor,
	enumitem,
	mathtools,
	bbold,
	ifthen,
	mdframed,
	multirow,
	subcaption,
	comment
}

\newcommand{\ts}{\hspace*{0.1em}} 

\newcommand{\N}{\mathbb{N}}                                     
\newcommand{\R}{\mathbb{R}}                                     
\newcommand{\innerprod}[2]{\left\langle #1,\, #2 \right\rangle} 
\newcommand{\dd}{\mathrm{d}}                                    
\providecommand{\norm}[1]{\left\lVert #1 \right\rVert}          

\newcommand\restr[2]{{\left.\kern-\nulldelimiterspace #1 \vphantom{\big|} \right|_{#2}}} 

\DeclareMathOperator{\domain}{dom}
\DeclareMathOperator{\range}{range}

\newcommand{\ink}{{k}} 
\newcommand{\outk}{{\ell}} 

\newcommand{\inspace}{\mathbb{X}} 
\newcommand{\inrkhs}{{H}} 
\newcommand{\outspace}{\mathbb{Y}} 
\newcommand{\outrkhs}{F} 

\newcommand{\infe}[1][]{
	\ifthenelse{\equal{#1}{}}{{\Phi}}{{\Phi}_{\scriptscriptstyle #1}}}

\newcommand{\outfe}[1][]{
	\ifthenelse{\equal{#1}{}}{{\Psi}}{{\Psi}_{\scriptscriptstyle #1}}}

\newcommand{\outreffe}[1][]{
	\ifthenelse{\equal{#1}{}}{{\Gamma}}{{\Gamma}_{\scriptscriptstyle #1}}}

\newcommand{\id}[1][]{
	\ifthenelse{\equal{#1}{}}{{I}}{{I}_{\scriptscriptstyle #1}}}

\newcommand{\idop}[1][]{
	\ifthenelse{\equal{#1}{}}{{\mathcal{I}}}{{\mathcal{I}}_{\scriptscriptstyle #1}}}

\newcommand{\ebdO}[1][]{
	\ifthenelse{\equal{#1}{}}{\mathcal{E}}{\mathcal{E}_{#1}}}
\newcommand{\pro}[1][]{
	\ifthenelse{\equal{#1}{}}{\mathcal{Q}}{\mathcal{Q}_{#1}}}

\newcommand{\cov}[1][]{C_{\scriptscriptstyle #1}} 
\newcommand{\ecov}[1][]{\widehat{C}_\mathit{\scriptscriptstyle #1}} 

\newcommand{\cinref}{\cov[\scriptstyle \inmeas]} 
\newcommand{\ecinref}{\ecov[\scriptstyle \inmeas]} 

\newcommand{\coutref}{\cov[\scriptstyle \outmeas]} 

\newcommand{\EEO}{\mathcal{U}_{Y \mid X}}
\newcommand{\eEEO}{\widehat{\mathcal{U}}_{Y \mid X}}

\newcommand{\PEO}{\mathcal{A}_{Y \mid X}}
\newcommand{\ePEO}{\widehat{\mathcal{A}}_{Y \mid X}}

\newcommand{\trans}{{\scriptscriptstyle \top}} 
\newcommand{\inv}{{\scriptscriptstyle  -1}} 

\newcommand{\ndat}{N} 
\newcommand{\nref}{M} 

\newcommand{\meas}{\mathbb{P}} 
\newcommand{\dens}{p} 

\newcommand{\me}[1][]{\mu_{#1}}  
\newcommand{\eme}[1][]{\widehat{\mu}_{#1}}  

\newcommand{\inmeas}{\rho} 

\newcommand{\outmeas}{\rho_y} 

\newcommand{\pf}[1][]{
   \ifthenelse{\equal{#1}{}}{\mathcal{P}}{\mathcal{P}_{#1}}}
\newcommand{\epf}[1][]{
   \ifthenelse{\equal{#1}{}}{\widehat{\mathcal{P}}}{\widehat{\mathcal{P}}_{#1}}}
\newcommand{\ko}[1][]{
   \ifthenelse{\equal{#1}{}}{\mathcal{K}}{\mathcal{K}_{#1}}}
\newcommand{\eko}[1][]{
   \ifthenelse{\equal{#1}{}}{\widehat{\mathcal{K}}}{\widehat{\mathcal{K}}_{#1}}}

\newcommand{\gram}[1][]{
	\ifthenelse{\equal{#1}{}}{{G}}{{G}_{\scriptscriptstyle #1}}}

\newcommand{\outgram}[1][]{
	\ifthenelse{\equal{#1}{}}{{L}}{{L}_{\scriptscriptstyle #1}}}

\newcommand{\ingram}[1][]{
	\ifthenelse{\equal{#1}{}}{{K}}{{K}_{\scriptscriptstyle #1}}}

\DeclareMathOperator{\mspan}{span}
\DeclareMathOperator{\diag}{diag}

\mathtoolsset{centercolon}

\allowdisplaybreaks

\newif\ifcomments
\newif\ifblind

\commentstrue 
\commentsfalse

\blindfalse

\ifblind
\newcommand{\unblindinfo}[2]{#2}
\else
\newcommand{\unblindinfo}[2]{#1}
\fi

\ifcomments 
\definecolor{krikcolor}{rgb}{0.0, 0.5, 0.7}
\newcommand{\ingmar}[1]{\textcolor{blue}{\textbf{Ingmar}: #1}}
\newcommand{\stefan}[1]{\textcolor{magenta}{\textbf{Stefan}: #1}}
\newcommand{\krik}[1]{\textcolor{krikcolor}{\textbf{Krik}: #1}}
\newcommand{\mattes}[1]{\textcolor{red}{\textbf{Mattes}: #1}}

\newenvironment{bingmar}
{\par\medskip \begingroup\color{blue} \ignorespaces \textbf{Ingmar:} }{\endgroup \medskip}

\newenvironment{bstefan}
{\par\medskip \begingroup\color{magenta} \ignorespaces \textbf{Stefan:}}{\endgroup \medskip}

\newenvironment{bkrik}
{\par\medskip \begingroup\color{krikcolor} \ignorespaces \textbf{Krik:}}{\endgroup \medskip}

\newenvironment{bmattes}
{\par\medskip \begingroup\color{red} \ignorespaces \textbf{Mattes:}}{\endgroup \medskip}
\else
\newcommand{\ingmar}[1]{{}}
\newcommand{\stefan}[1]{{}}
\newcommand{\krik}[1]{{}}
\newcommand{\mattes}[1]{{}}

\fi
\newtheorem{theorem}{Theorem}[section]
\newtheorem{corollary}[theorem]{Corollary}
\newtheorem{assumption}{Assumption}

\newtheorem{proposition}[theorem]{Proposition}

\theoremstyle{definition}

\newcommand{\mmu}{\mu_\mathbb{P}}
\newcommand{\emu}{\hat{\mu}_\mathbb{P}}
\newcommand{\eu}{\hat{u}}
\renewcommand{\Pr}[1]{\textrm{Pr} \left[ #1 \right]}

\hypersetup{draft}

\begin{document}
	%
	
	%
	
	\twocolumn[
	
	\aistatstitle{Kernel Conditional Density Operators}
	
	\aistatsauthor{ \unblindinfo{Ingmar Schuster \And Mattes Mollenhauer \And  Stefan Klus \And Krikamol Muandet}{Anonymous authors} }
	
	\aistatsaddress{ \unblindinfo{Zalando Research, Zalando SE \\	Berlin, Germany \And  Freie Universit\"at Berlin \\Berlin, Germany \And Freie Universit\"at Berlin \\Berlin, Germany   \And MPI 
		for Intelligent Systems\\	T\"ubingen, Germany}{Anonymous institution} } ]
	
	\begin{abstract}
		We introduce a novel conditional density estimation model termed the
		\emph{conditional density operator} (CDO). 
		It naturally captures multivariate, multimodal output densities and
		shows performance that is competitive with recent neural conditional density models and Gaussian processes.
		The proposed model is based on a novel approach to the
		reconstruction of
		probability densities from their kernel mean embeddings by drawing connections to estimation of Radon--Nikodym derivatives
		in the reproducing kernel Hilbert space (RKHS).
		We prove finite sample bounds for the estimation error in a standard
        density reconstruction scenario, independent of problem dimensionality.
		Interestingly, when a kernel is used that is also a
			probability density, the CDO allows us to
			both evaluate and sample the output density efficiently.
		We demonstrate the versatility and 
		performance of the proposed model on both synthetic and real-world data.
	\end{abstract}
	
\section{Introduction}
        
        Conditional density estimation is an essential task in
        statistics and machine learning \citep{Tsybakov08:INE,Di17}.  
        Popular techniques for estimating conditional densities
        include a kernel density estimator 
        \citep[KDE,][]{Tsybakov08:INE}, Gaussian process \citep[GP,][]{Wi06}, and deep
        neural networks \citep{Di17,papamakarios2017masked}. 
        While a KDE is simple to use, 
        it is known to suffer from the curse of
        dimensionality.
        While the GP is also a flexible model for conditional densities
        which enjoys a closed-form posterior thanks to the 
        Gaussianity assumption, approximate inference is often
        required to model  complex densities. 
        Lastly, deep neural networks have recently been used to model
        complex densities. 
        Despite a great representational power, they require large
        amount of training data and are prone to overfitting.

        The conditional mean embedding (CME) has emerged as an alternative
        kernel-based nonparametric representation for complex
        conditional distributions \citep{SHSF09,Song2013,MFSS16}.
        The CME can models complex distributions nonparametrically and can be
        estimated consistently from a finite sample. 
        It is mathematically elegant and is less prone to the curse of
        dimensionality, see, e.g., \cite{Tolstikhin17:MEK}. 
        However, one of the fundamental drawbacks of the CME is that a reconstruction
        of the associated conditional density becomes a non-trivial task. 
        To recover densities, a common approach is to  
        approximate them via a pre-image
        problem \citep{Kanagawa14:KPI,Song08:TDE}, which requires
        parametric assumptions on the densities. For sampling, kernel herding can be used \citep{Chen10:SKH}, which requires restrictive
        assumptions to ensure fast convergence.

	In this paper, we present a novel kernel-based supervised
        learning model for estimating conditional densities, the
        \emph{conditional density operator} (CDO). It has competitive performance with conditional density models based on deep neural networks~\citep{Di17}.
	To derive our model, we first present the problem of reconstructing a probability density from its associated kernel mean embedding~\citep{MFSS16,Smola07Hilbert} and
	connect it to the estimation of Radon--Nikodym derivatives.
	While this very general problem has been tackled before in similar scenarios~\citep{FSG13,QB13}, we provide a 
	characterization of conditions under which the density reconstruction as an inverse problem has
	a unique analytical solution.
	We show that in practical applications, 
	this statistical inverse problem can be solved conveniently
	using Tikhonov regularization~\citep{TA77,TikEtAl95}.
	Furthermore, we give finite sample concentration bounds for the 
	stochastic reconstruction error of the Tikhonov solution.
        When applied to conditional density estimation, our approach yields solutions that can capture multivariate, multimodal and non-Gaussian conditional densities and is not constrained by a homoscedastic noise assumption.
	This compares favorably with standard GPs and is on par with neural conditional density models \cite{Wi06,Di17}.
	In a set of experiments on toy and real-world data, we
        demonstrate that these properties lead to state-of-the-art
        results in conditional density estimation. 
        
        To summarize our contributions, we
        \emph{(i)} derive conditions under which a density can be
        reconstructed in the RKHS, \emph{(ii)} give a consistent
        estimator for the reconstructed density in the form of a
        statistical inverse problem, \emph{(iii)} provide
        dimensionality-independent finite sample error bounds for the
        estimation error of reconstructed densities, \emph{(iv)} introduce CDOs, a multivariate,
        multimodal kernel-based conditional density model.
	The rest of this paper is structured as follows:
	In Section~\ref{sec:Preliminaries and assumptions}, we state assumptions and introduce some preliminaries from the literature.
	Our main theoretical results are presented in Sections~\ref{sec:Density reconstruction and conditional density operators} and \ref{sec:cdo}, Section~\ref{sec:Related work} discusses related work.
	Experiments on a toy dataset, rough terrain estimation and traffic prediction are reported in Section~\ref{sec:Experiments}, while concluding remarks are presented in Section~\ref{sec:Conclusion}.
	
	\section{Preliminaries}
	\label{sec:Preliminaries and assumptions}

	\paragraph{Reproducing kernel Hilbert space (RKHS).}
     We only state the important facts here and collect related results in the supplementary material ~\cite[see also][Section 4.5]{StCh08}.
     We consider a measurable space $(\inspace, \Sigma)$, where $\inspace$ is a topological space
	endowed with the Borel $\sigma$-algebra $\Sigma$.
	Let $\ink \colon \inspace \times \inspace \rightarrow \R$ be a symmetric
	positive semidefinite kernel which induces an RKHS
	$\inrkhs = \overline{\mspan\{\ink(x, \cdot) \mid x \in \inspace \}}$, where
	the closure is with respect to 
	the inner product $\ink(x,x') = \innerprod{\phi(x)}{\phi(x')}_\inrkhs$. 
	Here, $\phi(x) := \ink(x, \cdot)$ is known as the \emph{canonical feature map}. 
     
\begin{assumption}[Separability] \label{ass:separability}
    The RKHS $\inrkhs$ is separable. Note that for a Polish space $\inspace$, 
    the RKHSs induced by a continuous kernel $\ink: \inspace \times \inspace \rightarrow \R$ is also separable~\citep[see][Lemma 4.33]{StCh08}.
    For a more general treatment of conditions implying separability, see~\cite{OwhadiScovel2017}.
\end{assumption}
 
	The \emph{reproducing property}
	$f(x) =  \innerprod{f}{\phi(x)}_\inrkhs$ holds for all $f \in \inrkhs$ and $x \in \inspace$.
	We fix a finite measure $\inmeas$ on $\inspace$ such that
    $\int_\inspace \norm{\phi(x)}_{\inrkhs}^2 \ts \dd \inmeas(x) < \infty$.
	Then the \emph{kernel mean embedding} 
    $\mu_\inmeas :=\int_{\inspace} \phi(x) \ts \dd \inmeas(x) \in \inrkhs$ of the measure
	$ \inmeas $ exists \citep{SFL11} and the (uncentered)
    \emph{kernel covariance operator}\footnote{Note that, technically, the term \emph{covariance operator} is misleading
    when $ \inmeas $ is not a probability measure.
    Since we will require $ \inmeas $
    to be finite, we will nevertheless use this term to reflect the standard definition.}
    $ C_\inmeas := \int_\inspace \phi(x) \otimes \phi(x) \ts \dd \inmeas(x) $ is well-defined as a positive self-adjoint Hilbert--Schmidt operator on $ \inrkhs $~\citep{Baker1973,Fukumizu04,MFSS16}. 
    Here, the map
    $ \phi(x) \otimes \phi(x) \colon \inrkhs \rightarrow \inrkhs $, given by $ f
    \mapsto \phi(x) \innerprod{f}{\phi(x)}_\inrkhs = \phi(x) \ts f(x)$ for
    all $ f \in \inrkhs $, is the rank-one \emph{tensor product
        operator}. 
    
    Whenever $ \inmeas $ is a
	probability measure, the kernel mean embedding admits the standard estimate $ \hat{\mu}_\inmeas := \nref^{-1} \sum_{i=1}^\nref \phi(x_i) $
	for i.i.d.\ samples $ (x_i)_{i=1}^\nref \sim \inmeas $.  
    For the covariance operator, we obtain the empirical estimate 
	$ \ecinref := \nref^{-1} \sum_{i=1}^\nref \phi(x_i) \otimes \phi(x_i) $ with samples as given above. Both $ \hat{\mu}_\inmeas $
	and $ \ecinref $ converge with $ \mathcal{O}(\nref^{-1/2})$ in probability
    in RKHS and Hilbert--Schmidt norm respectively~\citep{MFSS16}. 
	
	\paragraph{Inverse problems.}
    The general theory of inverse problems,
    pseudoinverse operators and 
    regularization
    has been well studied in the context of statistical learning over the last years~\citep{DeVitoEtAl2004,DeVitoEtAl2005,Caponnetto2007,Smale2007,DeVitoEtAl06}, we will therefore
    introduce these concepts only briefly.
    In general, the compact operator $\cinref$ can not be inverted on the whole 
    space $\inrkhs$.
    However, it admits a \emph{pseudoinverse}
    $\cinref^\dagger$, which is a (generally unbounded) operator with domain
    $\range(\cinref) + \range(\cinref)^\perp \subseteq \inrkhs$.
    Note that $\range(\cinref) + \range(\cinref)^\perp = \inrkhs$ if and only if
    $\range(\cinref)$ is a closed subspace, which is equivalent to $\inrkhs$ being finite dimensional.
    The minimum norm solution to the inverse problem 
    $\cinref u = f$ with known right-hand side $f \in \domain(\cinref^\dagger)$ is given by
    $u^\dagger := A^\dagger f$ and is unique, but solutions of larger norm can exist 
    in general.
    In practice,
    one can resort to the \emph{Tikhonov-regularized} solution 
    $u_\alpha := (\cinref + \alpha \idop[\inrkhs])^{-1} f$ (for a regularization 
    parameter $\alpha > 0$) to stabilize
    the problem against perturbed right-hand sides $\tilde{f}$ and ensure that the 
    solution is still well-defined even if $\tilde{f} \notin  \domain(\cinref^\dagger)$. 
    Note that as $\alpha \to 0$ we have $\norm{u^\dagger - u_\alpha}_\inrkhs \to 0$.
    Convergence rates for Tikhonov regularization schemes 
    have been derived in numerous settings depending on the problem and are usually connected to 
    rate of decay of the eigenvalues of $\cinref$. We refer the reader to the standard literature
    on inverse problems and regularization~\citep{TA77,EG96,TikEtAl95,EHN96} for details.
    	
        \paragraph{Conditional mean embedding.}
        The kernel mean embedding $\mu_{\rho}$ has been used
        extensively as a representation of the measure $\rho$ \citep{MFSS16}.
	We now extend this idea to conditional
        distributions~\citep{SHSF09,Gruen12,Song2013,MFSS16}.
	Note that \citep{SHSF09} formulates results in terms
	of (generally not existing) inverse operators under adequate regularity 
	assumptions.
	We use pseudoinverses instead of inverses, which aligns
	with the classical theory of inverse problems.
	Assume we have a topological output space $\outspace$ endowed with the Borel $\sigma$-algebra and a 
    positive semidefinite kernel
	$\outk \colon \outspace \times \outspace \rightarrow \R$ inducing a separable RKHS $\outrkhs$ 
    with feature map $\psi(y) := \outk(y, \cdot)$.
	All other assumptions we make for the space $\inspace$, its
	RKHS, and measures on $\inspace$ apply likewise for the output space
	$\outspace$ and associated objects. 
	We assume that random variables $X$ and $Y$
	with sample spaces $\inspace$ and $\outspace$ follow the joint distribution
	$\meas_{XY}$ with marginals $\meas_{X}, \meas_{Y}$ and induced
	conditional distribution $\meas_{Y \mid X}$. 
	Let $\cov[\mathit{YX}] := \int_\inspace \psi(y) \otimes \phi(x) \ts \dd \meas_{XY}(x,y)$  be the induced cross-covariance operator
	from $H$ to $F$ and $\cov[X]$ the covariance operator on $H$, respectively. 
	Then the \emph{conditional mean operator} (CMO) is defined as $\EEO = \cov[\mathit{YX}]\cov[X]^{\dagger} \colon \inrkhs \rightarrow \outrkhs $ and satisfies the equation $\me[\meas_y] = \EEO \me[\meas]$ for some distribution $\meas$ on $\inspace$,
	where $\meas_y(\cdot) = \int_{\inspace} \meas_{Y|X=x}(\cdot) \, \dd \meas(x)$~\citep{SHSF09,Song2013}. 
	In particular, if $\meas$ is the Dirac measure on $x' \in \inspace$,
	this yields $\me[\meas_{Y|X=x'}] = \EEO \ink(x',
        \cdot)$.\footnote{The latter is how the CMO is usually
          introduced, while $\me[\meas_y] = \EEO \me[\meas]$ is
          referred to as \emph{kernel sum rule} in the literature \citep{Fukumizu13:KBR}.}
	Note that the CMO is in general not a globally defined bounded operator. 
	It is defined pointwise as $\me[\meas_y] = \EEO \me[\meas] \in \outrkhs$ for
    $ \me[\meas] \in \domain \EEO$ under the condition 
    that $\mathbb{E}[g(Y) \mid X = \cdot \,] \in \inrkhs$ for all $g \in \outrkhs$. This requirement
	is examined in~\cite[Appendix A.1]{Fukumizu04}.
	In practical applications, the pseudoinverse $\cov[X]^{\dagger}$ is usually 
	replaced with its Tikhonov-regularized analogue, ensuring that $\EEO$ is 
	globally defined and bounded.
	
	\section{Density reconstruction from kernel embeddings}
	\label{sec:Density reconstruction and conditional density operators}
    
Our strategy to develop the CDO now is as follows. In this section, we will derive a method to reconstruct densities from their mean embeddings. This methodology we can then apply to \emph{conditional} mean embeddings and obtain the CDO as the aggregate of density reconstruction and conditonal mean operator.\\
    Assume we are given the mean embedding $\mmu$ of a target probability distribution $\mathbb{P}$.
    We now show how to reconstruct a Radon--Nikodym derivative $ \frac{\dd \mathbb{P}}{\dd \inmeas}$ with respect
    to a chosen finite positive \emph{reference measure} $\inmeas$ on $\inspace$
    that satisfies $\mathbb{P} \ll \inmeas$ and Assumptions~\ref{ass:rkhs_representative} and~\ref{ass:injectivity}.
    
    \begin{assumption}[RKHS representative] \label{ass:rkhs_representative}
        We assume that the $L_1(\inmeas)$ 
        equivalence class of the
        Radon--Nikodym derivative $ \frac{\dd \mathbb{P}}{\dd \inmeas}$ admits a representative
        which is an element of $H$. For simplicity, we will write  $ \frac{\dd \mathbb{P}}{\dd \inmeas} \in H$ for this representative.
    \end{assumption}

        We note that Assumption~\ref{ass:rkhs_representative} is not always satisfied in practice and
        is essentially a
        model assumption. However, the approximative qualities of RKHSs in terms of their ``size'' with respect to 
        other function spaces such as $C(\inspace)$ or $L_p(\inmeas)$, $p \in [1,\infty)$ are well examined --
        for a lot of kernels it can be shown that $\inrkhs$ is dense in these spaces~\cite{MXZ:Universal2006,StCh08,Sriperumbudur08injectivehilbert}.

    \begin{assumption}[Injective covariance operator] \label{ass:injectivity}
        The kernel covariance operator $\cinref$ exists 
        (i.e. $\int_\inspace \norm{\phi(x)}_{\inrkhs}^2 \ts \dd \inmeas(x) < \infty$) and is injective.
        Note that for example when $\ink$ is continuous on $\inspace \times \inspace$ and
        $\inmeas$ has full support on $\inspace$, the covariance operator is always injective~\citep{Fukumizu13:KBR}.
    \end{assumption}

    The theoretical background used in the derivation of the following results
    has appeared in a similar form in \cite{Fukumizu13:KBR}. We apply
    it in the context of density reconstruction and provide a formal mathematical setting 
    in terms of a statistical inverse problem which can elegantly be used in practice.
    The following result characterizes the Radon--Nikodym derivative $ \frac{\dd \meas}{\dd \inmeas} \in \inrkhs$
    as the solution of an inverse problem.

    \begin{proposition}[Radon--Nikodym derivatives]
        Let Assumption~\ref{ass:rkhs_representative} and~\ref{ass:injectivity} be satisfied.
        Then the inverse problem
        \begin{equation} \label{eq:main_result}
        \cinref u = \mmu, \quad u \in \inrkhs,
        \end{equation} has the unique solution 
        $u^{\dagger} := \cinref^\dagger \mmu =  \frac{\dd \mathbb{P}}{\dd \inmeas} \in \inrkhs$.
    \end{proposition}
    \begin{proof}
        We have
        $
        \cinref \tfrac{\dd \mathbb{P}}{\dd \inmeas}
        = \int\limits_\inspace \phi(x) \ts
        \tfrac{\dd \mathbb{P}}{\dd \inmeas} (x) \ts \dd \inmeas(x)
        = \int\limits_\inspace \phi(x) \ts \dd \mathbb{P}(x)
        = \mmu.
        $ Uniqueness of the solution follows directly since $\cinref$ is injective.
    \end{proof}

	Densities in the classical sense are Radon--Nikodym derivatives with respect to Lebesgue measure. This immediately gives the following special case.
	
    \begin{corollary}[Density reconstruction]
    	\label{cor:main_result} Let $\inspace \subseteq \R^d$ be compact and the kernel $\ink$ continuous.
    	Let $\inmeas$ be Lebesgue measure on $\inspace$ and $\mmu$ be the kernel mean embedding of a probability distribution $\mathbb{P}$ on $\inspace$. If  $\mathbb{P}$ admits a density and Assumption~\ref{ass:rkhs_representative} is satisfied, then the density is given by $\cinref^\dagger \mmu$.
    \end{corollary}
    Whenever we are given 
    an analytical mean embedding $\mmu$  in the setting of Corollary~\ref{cor:main_result},
    we can compute the unique solution $\cinref^\dagger \mmu$ and reconstruct the density of $\mathbb{P}$.
    In practice, we are usually given $\mmu$ in terms of an empirical estimate $\emu$, for
    example as an output of a mean embedding-based statistical model.
    We will now address the consistency and statistical details
    for the typical case that $\mmu$ is given in terms of its standard estimate $\emu$ and we can sample 
    from the reference measure $\inmeas$.  
    We emphasize that~\eqref{eq:main_result} can in theory be solved with any kind of numerical
    scheme for integral equations in the classical setting of inverse problems.
    
    \subsection{Consistency and convergence rate of the Tikhonov-regularized solution}
    \label{sec:Tikhonov solution}
    In practice, we cannot access $\cinref$ and $\mmu$ analytically.
    The idea is now to estimate $\cinref$ from an i.i.d. random sample.
    For the general case, this can be achieved by importance sampling.
    For the special case that 
    $\inmeas$ is the Lebesgue measure and $\inmeas(\inspace) =1 $, we can simply sample from it
    under the assumption that $\inspace$ is of convenient shape. We can then use the standard estimate for  $\ecinref$
    (see Section ~\ref{sec:Preliminaries and assumptions}). Additionally, we will assume
    for now that $\mmu$
    is also given in terms of its standard estimate $\emu$.
    Note that $\emu$ might instead be an estimate of
    a conditional mean embedding (as we will later consider) or an output of other
    model such as kernel Bayes rule \citep{FSG13}.
    Instead of computing the analytical density reconstruction $u^{\dagger} = \cinref^{\dagger} \mmu$, we construct an empirical
    estimate of $u^{\dagger}$ by defining the empirical Tikohonov-regularized solution 
    \begin{equation} \label{eq:empirical_solution}
    \eu := (\ecinref+\alpha\idop[\inrkhs])^{-1} \emu
    \end{equation}
    for a regularization parameter $\alpha > 0$.
    We examine this problem under the assumption that $\ecinref$
    is a standard empirical estimate based on
    $M$ i.i.d.\ $\inmeas$-samples and $\emu$ is estimated from $N$ i.i.d.\ $\mathbb{P}$-samples.
    Next, we show that the reconstruction error
    $\norm{u^{\dagger} - \eu}_\inrkhs$ vanishes in probability as $M,N \to \infty$ for an appropriately
    chosen positive regularization scheme $\alpha \to 0$ depending on sample sizes.
    We define the regularized solution 
    $u_\alpha := (\cinref + \alpha \idop[\inrkhs])^{-1} \mmu$ and
    decompose the total error:
    \begin{equation} \label{eq:error_decomposition}
    \norm{u^{\dagger} - \eu}_\inrkhs \leq \norm{ u^{\dagger} - u_\alpha }_\inrkhs + \norm{ u_\alpha - \eu}_\inrkhs.
    \end{equation}
    The first error term is deterministic and depends only on the analytical nature of the problem based on the decay of the eigenvalues of $\cinref$.

    The next result is based on a Hilbert space version of Hoeffding's inequality~\citep{Pinelis92,Pinelis94}
    and gives a general concentration bound for the estimation error term $ \norm{ u_\alpha - \eu}_\inrkhs $.
    
    \begin{proposition}[Finite sample bound of estimation error]
        \label{prop:stochastic_error}
        Let $\sup_x \sqrt{k(x,x)} = \sup_x \norm{\phi(x)}_\inrkhs = c < \infty$
        and $\alpha > 0$ be a fixed regularization parameter.
        Let $0 < a < 1/2$ and $0 < b < 1/2 $ be fixed constants. 
        If $\ecinref =  \nref^{-1} \sum_{i=1}^\nref \phi(x_i) \otimes \phi(x_i) $ with  $(x_i)_{i=1}^\nref \stackrel{\mathrm{i.i.d.}}{\sim} \inmeas$
        and $\emu =  \ndat^{-1} \sum_{j=1}^\ndat \phi(x'_j)$  with $(x'_j)_{j=1}^\ndat \stackrel{\mathrm{i.i.d.}}{\sim} \mathbb{P}$ and both
        sets of samples are independent,
        then we have
        \begin{equation}
        \begin{split}
        &\Pr{\norm{u_\alpha - \eu }_\inrkhs \leq \frac{\nref^{-2b}}{\alpha^2}(\norm{\mmu}_\inrkhs + \ndat^{-2a}) + 
            \frac{ \ndat^{-2a}}{\alpha}
        } \\ &\geq
        \left[ 1-2 \exp 
        \left(
        - \frac{\ndat^{1-2a}}{8 c^2} \right)
        \right]
        \left[ 1-2 \exp 
        \left(
        - \frac{\nref^{1-2b}}{8 c^4} \right)
        \right].
        \end{split}
        \end{equation}
    \end{proposition}
    
    The proof can be found in the supplementary material.
    We emphasize that the above error bound does not depend
    on the dimensionality of the data.
    By combining the convergence of the deterministic error and the convergence in probability given by
    Proposition~\ref{prop:stochastic_error}, we can obtain a
    regularization scheme which ensures that $\eu$ is a consistent
    estimate of $u^{\dagger}$. 
    \begin{corollary}[Consistency and regularization choice]\label{col:consistency}
        Let $\alpha = \alpha(\nref,\ndat)$ be a regularization
        scheme such that $\alpha(\nref,\ndat) \to 0$ 
        as well as 
        \begin{equation} \label{eq:consistency-scheme}
        \frac{\nref^{-2b}}{\alpha(\nref,\ndat)^2} \to 0 \quad \textrm{and} \quad
        \frac{ \ndat^{-2a}}{\alpha(\nref,\ndat)} \to 0
        \end{equation} as $\nref,\ndat \to \infty$.
        Then the empirical solution $\eu$ obtained from~\eqref{eq:empirical_solution}
        regularized with the scheme $\alpha(\nref,\ndat)$ converges
        in probability to the analytical solution
        $u^{\dagger}$.
        One such choice, given $c' \in (0, 1)$, is
        $\widetilde{\alpha}(\nref,\ndat) = \max(\nref^{-b},\ndat^{-2a})^{c'}$.
    \end{corollary}
    
    Small values for $c'$ imply larger bias and smaller variance (tighter bounds on the stochastic error), while large values for $c'$ imply smaller bias and larger variance.
    Note that Proposition~\ref{prop:stochastic_error} gives bounds only for the case that  $ \emu, \ecinref $ are given in
    terms of their standard empirical estimates.

        \section{Conditional density operators}
\label{sec:cdo}

        In this section, we use Corollary~\ref{cor:main_result} to
        define the \emph{conditional density operator} (CDO), which
        directly results in a conditional density for an output variable given an input variable or a distribution over the input variable. This is achieved by combining the density reconstruction method derived in the last section with conditional mean operators.\\ 
	Assume in what follows that we have fixed a finite positive reference measure $\outmeas$ on $\outspace$, such that $\coutref$ is a well-defined, injective, and positive self-adjoint Hilbert--Schmidt operator on $\outrkhs$.
	Moreover, densities on $\outspace$ are assumed to be Radon--Nikodym derivatives with respect to $\outmeas$ (we get densities in the usual sense if $\outmeas$ is Lebesgue measure).
	The following result is a direct consequence of Corollary~\ref{cor:main_result}.

	\begin{theorem}[Conditional density operator]
		\label{th:PEO}
		Assume $\meas_y(\cdot) = \int_{\inspace} \meas_{Y|X=x}(\cdot) \, \dd \meas(x)$ admits a density $p_y \in \outrkhs$ with respect to reference measure $\outmeas$, such that the assumptions
		of Corollary~\ref{cor:main_result} are satisfied. 
		Additionally assume that the conditional mean operator $\EEO = \cov[YX]\cov[X]^{\dagger}$ for $\meas_{Y|X}$
		exists and $\mmu \in \domain(\EEO$).
		Then 
        \begin{equation*}
        \PEO \me[\meas] := \cov[\outmeas]^{\dagger}\me[\meas_y] = \cov[\outmeas]^{\dagger}\EEO\me[\meas] = \cov[\outmeas]^{\dagger}\cov[YX]\cov[X]^{\dagger}\me[\meas] \in F
        \end{equation*}
		exists and satisfies $$ p_y {=}  \PEO \me[\meas].$$
		If $\meas$ is the Dirac measure on $x'$, this results in the density of $Y$ given $X=x'$ $$p_{Y|X=x'} {=}  \PEO \ink(x', \cdot).$$
	\end{theorem} 
	
\begin{figure*}
		\includegraphics[width=\textwidth]{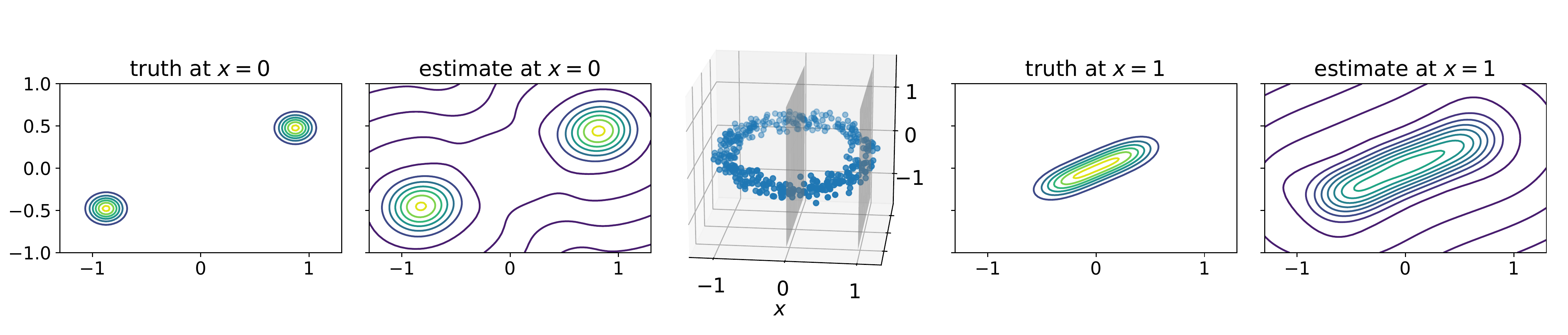}
		\caption{Cross sections of a donut shaped density and estimates using a CDO.}
		\label{fig:donut}
	\end{figure*}

	We call the operator $\PEO =
        \cov[\outmeas]^{\dagger}\cov[YX]\cov[X]^{\dagger}$
        mapping from $\inrkhs$ to $\outrkhs  $  the \emph{conditional density
          operator} (CDO).
        The CDO has several advantages over GPs, the mainstream kernel method for conditional density estimation \citep{Wi06}.
	In particular, it allows for density estimation in arbitrary output dimensions, unlike standard GPs, which estimate a $1$d density  \citep[see the literature on multi-output GPs for a remedy, e.g.][]{Al12,Bo05}.
	Moreover, multiple modes in the output can be captured by the CDO.
	Though this might be achieved with GP mixtures, the CDO allows for
        more flexibility as it requires no parametric assumptions on the mixture components.
        Heteroscedastic noise on the output is accounted for by standard CDOs, but nontrivial to include in GP models.
	 Interestingly, any output kernel that is also a probability density gives
         rise to CDOs where the output density can be both
         evaluated and sampled efficiently.
	 Also, CDOs allow uncertain inputs with any distribution, while closed form predictions for GPs  are only possible when the input uncertainty is Gaussian.
	Conditional densities estimated by a CDO are illustrated in Figure~\ref{fig:donut}; see Section~\ref{sec:donut} for a description of the data generating
	process.

    \subsection{Consistency of the conditional density operator}

    We can use the results from Section~\ref{sec:Tikhonov solution}
    to assess the consistency of the CDO.    
    The CDO is defined pointwise
    when the assumptions of Theorem~\ref{th:PEO} are satisfied.
    Analogously to the empirical inverse problem in Section~\ref{sec:Tikhonov solution},
    we replace the pseudoinverses of both $\cov[X]$ and $ \cov[\outmeas]$
    with their regularized inverses for the empirical version of the CDO.  
    From the (unbounded) analytical version 
    $\PEO =  \cov[\outmeas]^{\dagger}\cov[YX]\cov[X]^{\dagger}$, we obtain
    $\ePEO = (\ecov[\outmeas] + \alpha' \idop[\outrkhs]) ^{-1}
    \ecov[YX] (\ecov[X]+ \alpha \idop[\inrkhs])^{-1} $ which is a
    globally defined bounded operator.
    
	The proof of Proposition~\ref{prop:stochastic_error} in the supplementary material
    can directly be modified to see that whenever $\norm{\eme[\meas_y] - \me[\meas_y]}_\outrkhs \rightarrow 0$ 
    for a suitable regularization scheme $\alpha > 0$,
    we obtain a consistent regularized empirical solution 
    of the CDO when $\alpha' > 0$ is chosen appropriately.
    We will leave the statistical details to future work but want to emphasize that 
    the proof of Proposition~\ref{prop:stochastic_error} can also be used to obtain bounds
    for the conditional mean embedding by simply performing an additional composition
    with a cross-covariance operator. See~\cite{SHSF09,Fukumizu13:KBR,Fukumizu15:NBI} for
    asymptotic consistency results of the conditional mean embedding and appropriate regularization schemes.

	\subsection{Numerical representation of the conditional density operator}

	Assume that we have an i.i.d. sample $(x_{\scriptscriptstyle i},
        y_{\scriptscriptstyle i})_{i=1}^\ndat \sim \meas_{XY}$ such
        that the $\meas_{XY}$-induced conditional distribution
        $\meas_{Y \mid X}$ is the distribution of interest and another
        i.i.d. sample $(z_{\scriptscriptstyle i})_{i=1}^\nref 
	\sim  \outmeas$, where $\outmeas$ is the reference measure on $\outspace$ which we use to reconstruct the desired conditional density.
	The density over $\outspace$ induced by fixing the input at $x' \in \inspace$ is approximated as
	\begin{equation}
	\label{eq:ePEO point} 
	\ePEO \ink(x', \cdot) \approx \sum_{i=1}^{\nref}\beta_i \outk(z_{\scriptscriptstyle i}, \cdot)
	\end{equation}
	with $\beta = \nref^{-2} \ts (\outgram[Z]+  \alpha' \id[\nref])^{-2} \outgram[ZY] (\ingram[X] + \ndat \alpha \id[\ndat])^{-1} [\ink(x_{\scriptscriptstyle 1},x'), \dots, \ink(x_{\scriptscriptstyle \ndat}, x')]^\trans \in \R^\nref$, where we use the kernel matrices $\ingram[X] = \left[\ink(x_i,x_j) \right]_{ij} \in \R^{\ndat \times \ndat}$, 
	$\outgram[Y] = \left[\outk(y_i,y_j) \right]_{ij} \in \R^{\nref \times \nref}$ as well as
	$\outgram[ZY] = \left[\outk(z_i,y_j) \right]_{ij} \in \R^{\nref \times \ndat}$ and
	the corresponding identity matrices $\id_\ndat \in \R^{\ndat \times \ndat}, \id_\nref \in \R^{\nref \times \nref}$.
	If one is interested in the marginal distribution of $Y$ when integrating out $x' \sim \meas$, the $\ink(x_{\scriptscriptstyle j}, x')$ are replaced by $\me[\meas](x_{\scriptscriptstyle j})$ in the expression for $\beta$.
	The derivation of the representation in~\eqref{eq:ePEO point} 
	builds upon a similar derivation of the conditional mean embedding estimate 
	and can be found in the supplementary material.
	Detailed convergence rates and error bounds for this empirical estimate are 
	beyond the scope of this paper.

	\section{Related work}
	\label{sec:Related work}
	
	Finding the pre-image of a feature vector in the RKHS is a classical problem in kernel methods
	\citep{Kwok03:PP,Bakir04:PI}. 
	In this work, our goal is to reconstruct a density $p$ from the kernel
	mean embedding $\mu_{\meas}$ of some distribution $\meas$. 
	There exist two popular approaches in the literature for recovering information from $\mu_{\meas}$, namely distributional pre-image learning \citep{Song08:TDE,Kanagawa14:KPI} and
	kernel herding \citep{Chen10:SKH}.
        Given an empirical kernel mean $\hat{\mu}_{\meas}$, the idea
        of the former is to pick a
	family of densities $\mathcal{P}=\{p_{\theta}\,:\,\theta\in\Theta\}$ and then
	find $\theta^* =
	\arg\min_{\theta\in\Theta}\|\hat{\mu}_{\meas} -
	\mu_{p_{\theta}}\|^2_H$ \citep{Song08:TDE}. 
	The drawback of this approach is the parametric
	assumptions on the family of densities $\mathcal{P}$.
	Moreover, it requires solving a constrainted non-convex optimization problem. A related approach is that of using \citep{TonChaTehSej2019}, which suggests to use conditional means as an input for a neural density model.
	On the other hand, our method provides an analytic solution for conditional densities which only requires that $\meas$ is absolutely
	continuous with respect to the reference measure $\inmeas$.
        Alternatively, kernel herding aims to greedily generate a representative
	set of $ T $ pseudo-samples from $\meas$ in a deterministic fashion using the estimate
	$\hat{\mu}_{\meas}$ \citep{Chen10:SKH}. 
	The advantage of herding is an integration error of order
	$\mathcal{O}(T^{-1})$ under some assumptions. 
	Similarly, our method gives rise to a probability density from which
	random samples can be easily generated. 
	Note that while our work also relates to the literature of 
	kernel-based density ratio estimation \citep{Kanamori12:SAK,QB13}, our goal is not to
	estimate a density ratio.
	Furthermore, unlike previous work, we provide a rigorous treatment of the error bounds of our estimates and good choices for regularization constants.
	Lastly, the kernel mean embedding has recently been applied to 
	fit high-dimensional \emph{implicit} density models such as
	generative adversarial networks (GANs)
	\citep{dziugaite2015training,li2015generative, li2017mmd} and
	autoencoders \citep{tolstikhin2018wasserstein}. 
	It would also be interesting to extend our results to this
        area of research.

	Classical methods for (conditional) density estimation \citep{BH01,HRL04} are known to suffer from slow
	convergence in high dimensions, see, e.g.,
	\citet[Chap. 1]{Tsybakov08:INE}.
	Some methods propose estimators that are similar to the CDO, although not making use of RKHS arguments and not proving consistency \citep{BH01}.
	An advantage of the CDO is that it is less prone to the curse of
	dimensionality.
	Concretely, the convergence rate of Proposition~\ref{prop:stochastic_error}
	and regularizing scheme from Corollary~\ref{col:consistency}
	do not depend on the problem dimension.
	Nevertheless, it might
	affect the deterministic error which could converge arbitrarily slowly,
	see, e.g., \citet{Tolstikhin17:MEK}.
	Neural density models can also scale gracefully with increasing dimensions, as demonstrated empirically especially in the image generation domain \citep{KD18,Di17}.
	However, little theory exists to confirm this observation and understand under which conditions on the problem and the network architecture it applies.
	Standard neural density models can easily be extended to
        include conditioning on an input variable. However,
        conditioning on a distribution over the input variable is
        non-trivial, unlike in the CDO setting.

In the RKHS setting, infinite dimensional exponential families (IDEF) and their conditional extension, kernel conditional exponential families (KCEF) assume the log-likelihood of a (conditional) density to be an RKHS function 
 \citep{SFGHK17,AG18}.
	Fitting such a model is solved by using an optimization approach, while CDOs allow closed form solutions.
	Furthermore, CDOs allow for trivial normalization of the estimated densities unlike kernel exponential family approaches.
	Sampling from IDEF and KCEF approximations requires MCMC techniques rather than ordinary Monte Carlo as with our approach.
	Sampling is necessary to estimate predictive mean and variance in IDEF models, while closed form expressions exist for CDOs, see  \ref{sec:mean variance closed form}.

	\section{Experiments}
	\label{sec:Experiments}	
	
	In this section, we report results on one toy and two real-world datasets, showing competitive performance of the CDO in conditional density estimation tasks in comparison to recent state-of-the-art approaches.
	We use a computational trick for large datasets which is described, along with a trick for high-dimensional output spaces, in the supplementary material~\ref{sec:computational tricks}. For regularization of the CDO in the experiments that follow, we always use Corollary~\ref{col:consistency} with $c' = 0.99999$.  Neural density models used as baseline methods where implemented in PyTorch and optimized using ADAM \citep{Ki14} with a learning rate tuned to achive the best training log likelihood. Hidden layers contained $100$ ReLUs each. The optimal number of hidden layers differed and is stated per experiment.
	
	\subsection{Gaussian donut}
	\label{sec:donut}
	\begin{figure}
		\centering
		\includegraphics[width=0.95\linewidth]{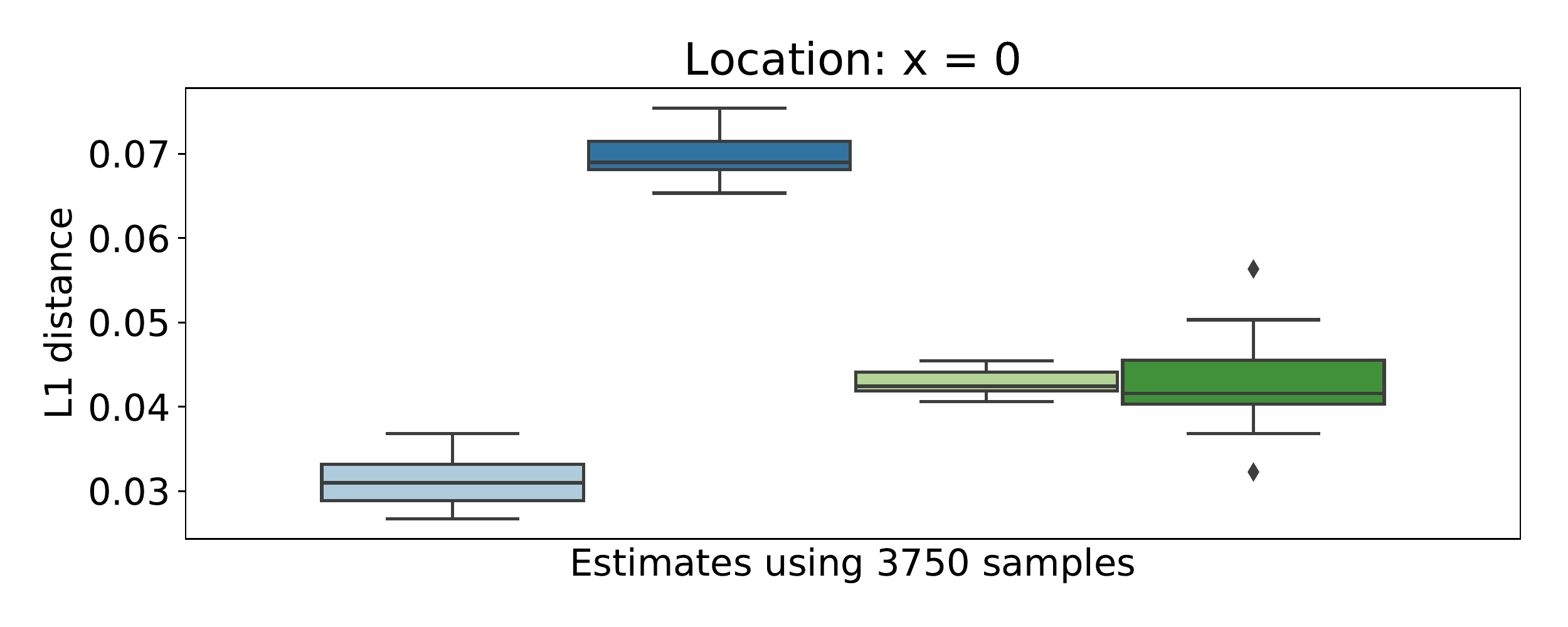}\\
		\includegraphics[width=0.95\linewidth]{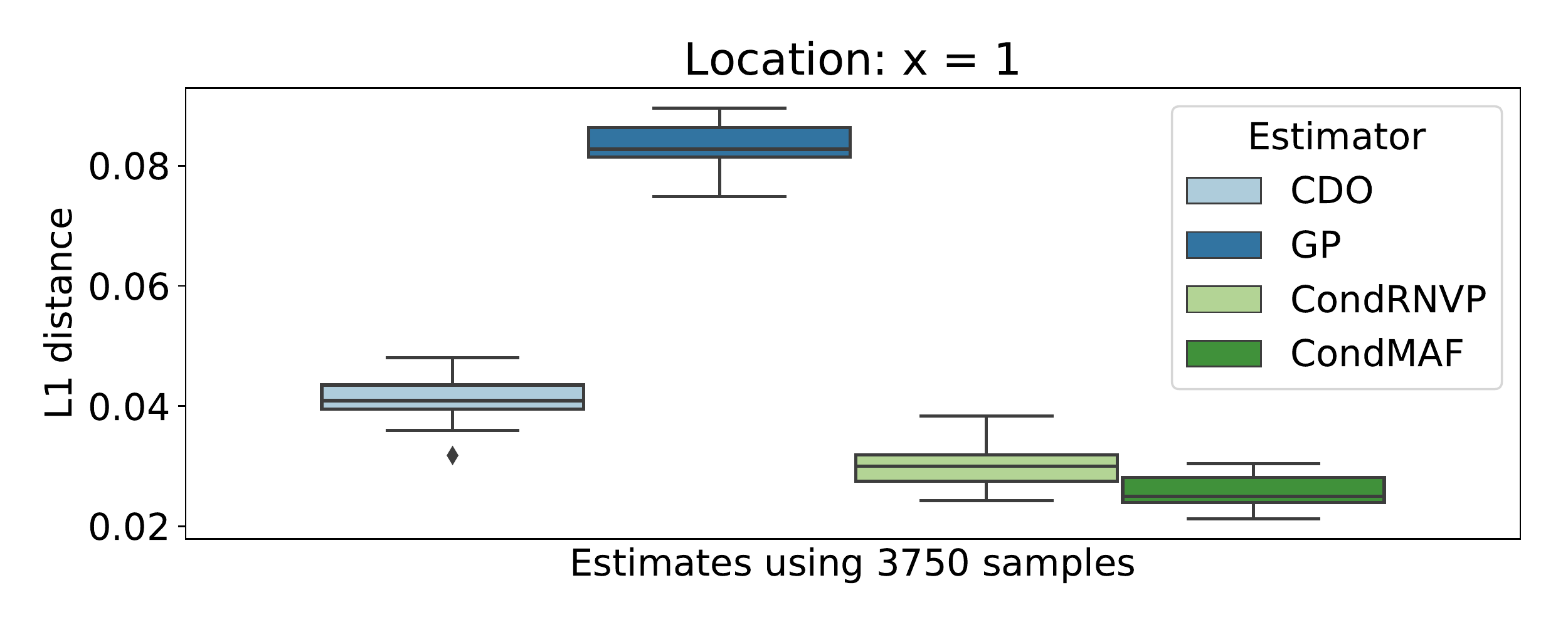}
		\caption{Errors of conditional density estimation for the Gaussian donut in $L_1(\outmeas)$-norm.}
		
		\label{fig:Donut norm}
	\end{figure}
	For this toy example, a unit circle in the $(x,y)$ plane is embedded into a $3$D ambient space and slightly rotated around the $y$ axis, then we pick $50$ equidistant points on this circle.
	Each of the points is the mean of an isotropic Gaussian distribution, and each mean has equal probability, giving rise to a mixture that we call a Gaussian donut.
	We draw $50$ samples from the isotropic Gaussian noise distribution per mean to form the training data for a CDO that estimates the density on $y,z$ coordinates given $x$.
	The reference measure $\outmeas$ is taken to be the uniform distribution on a zero-centered square with side length $4$.
	See Figure~\ref{fig:donut} for the ground truth density and CDO estimate at $x$ equal to $0$ and $1$, respectively. 
	We report numerical errors in density approximation in $L_1(\outmeas)$-norm, i.e., $\|\widehat{u} - \dens_{Y,Z\mid X=x}\|_{L_1}$ in Figure~\ref{fig:Donut norm}. Input and output kernels where Laplace and Gaussian with lengthscale resulting from the median heuristic \citep{garreau2017large}.
	The uniform reference measure is represented by a regular grid of $\nref = \lfloor\sqrt{\ndat}\rfloor^2$ points.
	The procedure is repeated $10$ times for different random seeds.
	One comparison method are GPs on each output dimension independently with a Gaussian kernel and lengthscale optimized for highest marginal likelihood with GPyTorch \citep{GPW18}. Furthermore, we use conditional versions of RealNVP and masked autoregressive flows (MAF) with $5$ hidden layers, see \cite{Di17,papamakarios2017masked}. The KCEF method \citep{AG18} could not be used here because it yields unnormalized density estimates we could not get reliable estimates of the normalizing constant.
	See Figure~\ref{fig:Donut norm} for a plot of the $L_1$ error, which shows that the CDO provides competitive conditional density estimates on this simple dataset.
	
	\subsection{Rough terrain reconstruction}
	Rough terrain reconstruction is used in robotics and navigation \citep{Ha10,Gi10}.
	Given measurements of longitude, latitude, and altitude, the task is to estimate the altitude for unobserved coordinates on a map.
	We reproduce an experiment from \cite{Er18}, considering around $23$ million non-uniformly sampled measurements of Mount
	St. Helens, binned into a $120 \times 117$ grid.
	We randomly chose $80\%$ of the data as training, the rest as test data.
	We fit an exact Gaussian Process by optimizing the length scale of a Gaussian kernel with respect to marginal likelihood of the training data and compute the scaled mean absolute error (SMAE) for the test locations.
	Furthermore, we fit neural conditional density models based on RealNVP and MAF \citep{Di17,papamakarios2017masked}, where $10$ hidden layers gave the best training log likelihood.
	For the conditional density operator, we pick a Gaussian kernel for input and output domains.
	The input length scale is chosen using the median heuristic \citep{garreau2017large}.
	The output domain is chosen as an interval based on the minimum and maximum of the training output data, with a uniform reference measure represented by equidistant grid points,
	the output length scale based on the distance between adjacent grid points.
	See Table~\ref{tab:rough terrain} for a summary of the SMAEs  reached by each method.
	The result again suggests that our method is competitive with other
	kernel-based learning algorithms and recent neural density models.
	We conjecture that added flexibility is a reason for for the better performance compared to a GP.
	While the output distribution of the GP is a Gaussian, in the CDO used here it is a mixture of Gaussians.
	A related possibility is that we use a homoscedastic likelihood in the GP, leading to a certain minimum amount of assumed noise, while the CDO does not do this.
	
		\begin{table}
		\caption{Test Set SMAEs rough terrain}
		\label{tab:rough terrain}
		\centering
		
		\begin{tabular}{lc}
			\toprule
			{Estimator}  		& {SMAE}\\
			\midrule
			\textbf{CDO}&  $0.0269\pm0.0006$\\
			\textbf{GP}&  $0.0358\pm0.0006$\\
			\textbf{Cond. Real NVP} & $0.0373 \pm 0.0380$\\
			\textbf{Cond. MAF} &$0.0309 \pm 0.0395$\\
			\bottomrule
		\end{tabular}
	\end{table}
	
	\subsection{Traffic density prediction from time features}
	\begin{figure*}
		
		\includegraphics[width=\linewidth]{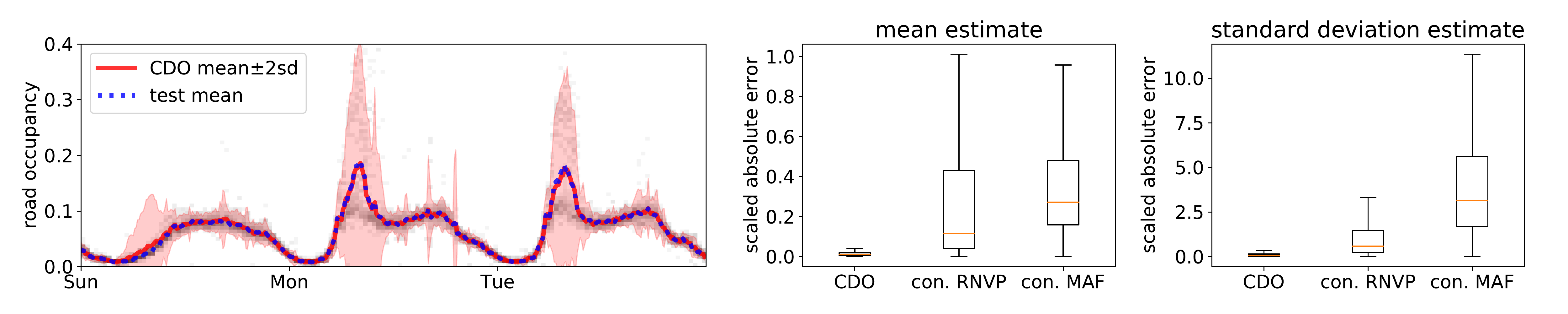}
		\caption{Road occupancy prediction experiment. \emph{Left:} Histogram of test data for three days in black, test data mean and prediction.  \emph{Right:} Boxplots of scaled absolute errors with respect to test data.}
		
		\label{fig:traffic}
	\end{figure*}
	
	In this experiment, we predict the occupancy rate of different locations on freeways in the San Francisco bay area based on a given day of week and time of day.\!\footnote{Detailed description in \cite{Cu11}, data available at \href{https://archive.ics.uci.edu/ml/datasets/PEMS-SF}{https://archive.ics.uci.edu/ml/datasets/PEMS-SF}.}
	The occupancy rate is encoded as a number between $0$ and $1$ for $963$ different locations.
	The measurements are sampled every 10 minutes, resulting in $144$ measurements per day (i.e., times of day).
	See Figure~\ref{fig:traffic} for example histograms at one particular location.\\
	In the training dataset, each day of week occured $32$ times (discarding measurements to get a balanced dataset), resulting in $32 \times 144 \times 7 = 32\,256$ input-output pairs. In the test set, each day of week occured $20$ times.
	The task is to get a predictive density for the locations occupancy given time of day and day of week as inputs. \\
	We fit a conditional density operator using Gaussian kernels on the output  and Laplacian kernels on
	the input domain.
	Laplacians are chosen because they result in smoother estimates,
	while Gaussians showed more oscillations for the output density estimates.
	Samples for the uniform reference measure on the output domain are taken to be a regular grid between minimum and maximum occuring values.
	Bandwidth for both kernels is chosen based on the median heuristic.
	For comparison, we use both RealNVP  and MAF deep neural networks \citep{Di17,papamakarios2017masked}, where $5$ hidden layers gave the best training log likelihood. 
	We  estimate the expectation (w.r.t.\ model predictive distribution) of the absolute error when estimating test set mean and variance, i.e., scaled mean absolute error (SMAE), and its standard deviation. Mean and variance are chosen because closed form estimates of these exist under the CDO.
	As this is not the case for the neural models, we draw $2000$ samples for
	estimation.
	Even though the dataset is rather large, the CDO can be fitted in under one minute on a modern laptop using a scheme outlined in~\ref{sec:Large datasets trick using factorization of the joint probability}. Because we could not adapt this scheme to KCEF \citep{AG18}, it was impossible to fit this alternative kernel conditional density model, because memory requirements could not be satisfied even on a large compute server. Errors are summarized in Table~\ref{tab:traffic} and plotted in Figure~\ref{fig:traffic}.
	Clearly, our CDO outperforms the neural models. While we also fitted GPs using the GPyTorch package, the errors where huge because this problem necessitates heteroscedastic likelihood noise, which is unavailable in currently maintained GP packages.

	\begin{table}
		\caption{Test Set SMAEs Road Occupancy Data}
		\label{tab:traffic}
		\centering

		\begin{tabular}{lcc}
			\toprule
			{ } 		& {mean} & {sd}\\
			\midrule
			\textbf{CDO}&  $0.02 \pm 0.05$ & $0.36 \pm 1.21$\\
			\textbf{Cond. Real NVP} & $0.32\pm 0.41$ &$1.19 \pm 1.65$\\
			\textbf{Cond. MAF} &$0.52 \pm 0.81$& $4.50 \pm 4.40$\\
			\bottomrule
		\end{tabular}

	\end{table}

	\section{Conclusion}
	\label{sec:Conclusion}
	
	In this paper, we show that the reconstruction of densities from kernel
	mean embeddings can be formulated as an inverse problem under some 
	regularity assumptions.
	In particular, we draw connections to the estimation of Radon--Nikodym derivatives with respect to a Lebesgue reference measure, for which the solution is shown to be unique.
	We prove that the popular Tikhonov approach to solving the inverse problem is consistent and allows for finite sample bounds on the estimation error independent of the
	dimensionality of the data.
	However, we want to point out that the proposed Tikhonov scheme is only one possible approach for finding a solution.
	We focus on the conditional density operator as an straight forward application of the density reconstruction result.
	The CDO is closely related to the conditional mean embedding, can model multivariate, multimodal conditional distributions and performs competitively in our experiments.\\
	In future work, numerical routines for scaling the method up to even larger datasets will be of interest.
	One way to do this might be conjugate gradient algorithms and making use of Toeplitz and Kronecker structure in the kernels, as recently done in fitting GPs \citep{GPW18,WN15}.
	Theoretical avenues to take might be to find rigorously justified ways of choosing good kernels and kernel parameters.
	
	\subsubsection*{Acknowledgments}

	\unblindinfo{
	We would like to thank Kashif Rasul for providing the conditional RealNVP implementation for the traffic dataset and Ilja Klebanov and Tim Sullivan for helpful discussions and pointing out relevant references.
	 Partially funded by the Deutsche Forschungsgemeinschaft (DFG, German Research Foundation) under Germanys Excellence Strategy -- MATH+: The Berlin Mathematics Research Center, EXC-2046/1 – project ID: 390685689.}{WITHHELD FOR BLIND REVIEW}
	\bibliographystyle{abbrvnat}
	\bibliography{library}	
\newpage
\onecolumn
	\appendix
		\section{Supplementary material}

	\subsection{Proofs for results in the main text}
	
	Here we provide the proofs which were omitted in the main text due to the page limitation.

	\begin{proof}[Proof of Proposition~\ref{prop:stochastic_error}]
		Since we assume 
		$\sup_x \sqrt{k(x,x)} = \sup_x \norm{\phi(x)}_H = c < \infty$,
		we can apply a Hilbert space version of Hoeffding's inequality~\citep{Pinelis92,Pinelis94}
		to obtain the following concentration bounds \citep{RBD10,Schneider16}.
		For every $\epsilon_1, \epsilon_2 > 0$, we have
		\begin{equation} \label{eq:hoeffding_mu}
		\Pr{ \norm{\mmu - \emu}_H \leq \epsilon_1} \geq 1 - 2 \exp 
		\left(
		- \frac{\ndat \epsilon_1^2}{8 c^2}
		\right)
		\end{equation}
		as well as
		\begin{equation} \label{eq:hoeffding_cov}
		\Pr{ \norm{\cinref - \ecinref}\leq \epsilon_2} \geq 1 - 2 \exp 
		\left(
		- \frac{\nref \epsilon_2^2}{8 c^4}
		\right),
		\end{equation}
		where the estimates are based on $ \ndat $ and $ \nref $ i.i.d.\ samples from $\mathbb{P}$ 
		and $ \inmeas $, respectively. 
        Note that the bound~\eqref{eq:hoeffding_cov} is first obtained in Hilbert--Schmidt norm
        and based on the fact the the operator norm is aways dominated by the Hilbert--Schmidt norm.
        We assume that~\eqref{eq:hoeffding_mu} and~\eqref{eq:hoeffding_cov} hold independently.
		We remark that every alternative concentration bound for the above
		estimation errors can be used in the same way below, leading to analogue results.
		
		For every fixed $\alpha > 0$ and corresponding solution to the regularized empirical 
		and analytical problem ($ \eu = (\ecinref  + \alpha \idop[\inrkhs])^{-1} \emu$ and $u_\alpha = (\cinref + \alpha \idop[\inrkhs])^{-1} \mmu$, respectively), we have
		\begin{align*}
		\norm{\eu - u_\alpha}_H
		&= \norm{
			(\ecinref  + \alpha \idop[\inrkhs])^{-1} \emu - (\cinref  + \alpha \idop[\inrkhs])^{-1} \mmu
		}_H
		\\ &\leq 
		\underbrace{
			\norm{(\ecinref  + \alpha \idop[\inrkhs])^{-1} \emu - (\cinref  + \alpha \idop[\inrkhs])^{-1} \emu}_H 
		}_{(\star)} \\&+
		\underbrace{
			\norm{(\cinref  + \alpha \idop[\inrkhs])^{-1} \emu - (\cinref  + \alpha \idop[\inrkhs])^{-1} \mmu}_H
		}_{(\star \star)}.
		\end{align*}
		Using the fact that $\ecinref$ and $\cinref$ are both self-adjoint and positive,
		we have $\norm{(\ecinref + \alpha \idop[\inrkhs])^{-1}}  \leq \frac{1}{\alpha}$ as well as 
		$\norm{(\cinref + \alpha \idop[\inrkhs])^{-1}}  \leq \frac{1}{\alpha}$. Together with the identity 
		$ A^{-1} - B^{-1} = A^{-1}(B-A)B^{-1} $ for all bounded  linear operators
		$A$ and $B$, we get
		\begin{equation*}
		\norm{(\ecinref + \alpha \idop[\inrkhs])^{-1}  
			- (\cinref + \alpha \idop[\inrkhs])^{-1}}  \leq \frac{1}{\alpha^2} \norm{\ecinref  - \cinref}.
		\end{equation*}
		We use the above inequality to bound the term $(\star)$ as
		\begin{align*} (\star) 
		\leq \frac{1}{\alpha^2} \norm{\ecinref  - \cinref} \norm{\emu}_H
		\leq \frac{\epsilon_2}{\alpha^2} \left(\norm{\mmu}_H + \epsilon_1\right)
		\end{align*} 
		and the term $(\star \star)$ as
		\begin{align*} 
		(\star \star) \leq \norm{(\cinref + \alpha \idop[\inrkhs])^{-1}} \norm{\mmu - \emu}_H \leq \frac{\epsilon_1}{\alpha}.
		\end{align*}
		Both bounds hold simultaneously
		with probability of at least 
		\begin{equation*}
		\left[ 1-2 \exp 
		\left(
		- \frac{\ndat \epsilon_1^2}{8 c^2} \right)
		\right]
		\left[ 1-2 \exp 
		\left(
		- \frac{\nref \epsilon_2^2}{8 c^4} \right)
		\right]
		\end{equation*}
		as given by~\eqref{eq:hoeffding_mu} and~\eqref{eq:hoeffding_cov}. Note that this
		implies
		$\norm{ \eu - u_\alpha }_H \leq \frac{\epsilon_2}{\alpha^2}(\norm{\mmu}_H + \epsilon_1) + \frac{\epsilon_1}{\alpha} $ with
		the same probability by the inequalities above.
		We now express the resulting bound in terms of sample sizes $\nref$ and $\ndat$.
		Since the above concentation bounds hold for arbitrary $\epsilon_1, \epsilon_2 > 0$,
		we can fix coefficients $0 < a < 1/2$ and $0 < b < 1/2 $
		and set $\epsilon_1 := \ndat^{-a}$ and $\epsilon_2 := \nref^{-b}$,
		resulting in 
		\begin{equation*}
		\norm{ \eu - u_\alpha}_H \leq \frac{\nref^{-2b}}{\alpha^2}(\norm{\mmu}_H + \ndat^{-2a}) + 
		\frac{ \ndat^{-2a}}{\alpha}.
		\end{equation*} with a probability of at least
		\begin{equation*}
		\left[ 1-2 \exp 
		\left(
		- \frac{\ndat^{1-2a}}{8 c^2} \right)
		\right]
		\left[ 1-2 \exp 
		\left(
		- \frac{\nref^{1-2b}}{8 c^4} \right)
		\right]. \qedhere
		\end{equation*}
	\end{proof}

	\subsection{Numerical representation of $\ePEO$ based on training data}
	\label{sec:Numerical representation of empirical CDO based on training data}
	
	In what follows, we derive a closed form expression for $\ePEO = (\ecov[Z] + \alpha'\idop[\outrkhs])^{-1} \eEEO$ which can
	be approximated numerically given a fixed input $x' \in \inspace$.
	
	We adopt the so-called \emph{feature matrix notation} \cite{MFSS16,SHSF09}
	and define $\infe = [\ink(x_{\scriptscriptstyle 1},\cdot), \dots, \ink(x_{\scriptscriptstyle \ndat}, \cdot)]$ and
	$\outfe = [\outk(y_{\scriptscriptstyle 1},\cdot), \dots, \outk(y_{\scriptscriptstyle \ndat}, \cdot)]$. 
	We express the Gram matrix for $X$ as $\ingram[X] = \infe^\trans \infe$.
	Then we have the standard estimates $\cov[YX] \approx \ecov[YX] = \ndat^{-1} \outfe \infe^\trans$
	and $\ecov[X] = \ndat^{-1} \infe \infe^\trans$.
	Assume additionally that we have drawn samples from  $\outmeas$ and let $\outreffe = [\outk(z_{\scriptscriptstyle 1},\cdot), \dots, \outk(z_{\scriptscriptstyle \nref}, \cdot)]$ for $(z_{\scriptscriptstyle i})_{i=1}^\nref \stackrel{\mathrm{i.i.d.}}{\sim}  \outmeas$.
	Let $Z$ be a $\outmeas$-distributed random variable.
	This implies $\coutref  \approx \ecov[Z] = \nref^{-1} \outreffe \outreffe^\trans$.
	
	It is well known
	that $\nref^{-1} \outgram[Z] = \nref^{-1} \outreffe^\trans \outreffe \in \R^{\nref 
		\times \nref}$ and the empirical covariance operator $\ecov[Z]$ share
	the same nonzero eigenvalues and their eigenvectors/eigenfunctions can
	be related. This fact has been examined a lot in various scenarios, see for
	example~\cite{Shawe-TaylorEtAl2002,RBD10}. In particular,
	we have the relation
	\begin{equation*}
	M^{-1} \outgram[Z]	= V \Lambda V^\trans \quad \Leftrightarrow \quad 
	\ecov[Z] = 
	\sum_{i = 1}^{r} \lambda_i \; 
	( \lambda_i^{-1/2}\outreffe v_i) \otimes ( \lambda_i^{-1/2}\outreffe v_i)
	= (\outreffe V \Lambda^{-1/2}) \ts 
	\Lambda \ts (\outreffe V \Lambda^{-1/2})^\trans,
	\end{equation*}
	where $\Lambda = \diag(\lambda_1, \dots, \lambda_r, 0, \dots, 0) 
	\in \R^{\nref \times \nref}$ contains the $r \leq \nref$ 
	nonzero eigenvalues $\lambda_i$
	of $M^{-1} \outgram[Z]$
	corresponding to unit norm eigenvectors 
	$v_i \in \R^\nref$ and
	$\Lambda^{-1/2} = 
	\diag(\lambda_1^{-1/2}, \dots, \lambda_r^{-1/2}, 0, \dots, 0)$.
	
	Hence, the $\outrkhs$-normalized eigenfunctions of $\ecov[Z]$ are given by
	$\lambda_i^{-1/2}\outreffe v_i 
	=\lambda_i^{-1/2} \ts \sum_{j=1}^{\nref} v_i^{(j)} \outk(z_j,\cdot)$.
	Note that $\outrkhs = \mspan\outreffe \oplus (\mspan\outreffe)^\perp$.
	For a closed subspace $U \subseteq \outrkhs$,
	let $P_{U}$ denote the orthogonal projection operator
	onto $U$.
	Based on the eigendecomposition of $\ecov[Z]$,
	we naturally have
	\begin{equation*}
	\ecov[Z] + \alpha'\idop[\outrkhs] =
	(\outreffe V \Lambda^{-1/2}) \ts 
	(\Lambda + \alpha'\id[\nref] ) \ts (\outreffe V \Lambda^{-1/2})^\trans +  \alpha'P_{(\mspan\outreffe)^{\perp}}
	\end{equation*}
	for any fixed regularization parameter $\alpha' > 0$.
	As an immediate consequence, we obtain
	\begin{align*}
	(\ecov[Z] + \alpha'\idop[\outrkhs])^{-1} &=
	(\outreffe V \Lambda^{-1/2}) \ts 
	(\Lambda + \alpha'\id[\nref] )^{-1} \ts (\outreffe V \Lambda^{-1/2})^\trans + \alpha'^{-1} P_{(\mspan\outreffe)^{\perp}}
	\\
	&= \outreffe V (\Lambda^{-1/2} \Lambda^{-1/2})  \ts 
	(\Lambda + \alpha'\id[\nref] )^{-1} V^\trans \ts \outreffe^\trans + \alpha'^{-1} P_{(\mspan\outreffe)^{\perp}}
	\\
	&= \outreffe V (\Lambda^{-1/2} \Lambda^{-1/2})  \ts V^\trans V \ts
	(\Lambda + \alpha'\id[\nref] )^{-1} V^\trans \ts \outreffe^\trans + \alpha'^{-1} P_{(\mspan\outreffe)^{\perp}}
	\\
	& = \nref^{-2} \ts \outreffe \outgram[Z]^\dagger \ts 
	(\outgram[Z] + \alpha'\id[\nref] )^{-1} \ts \outreffe^\trans + \alpha'^{-1} P_{(\mspan\outreffe)^{\perp}},
	\end{align*}
	Where we use that $\Lambda^{-1/2}$ and $(\Lambda + \alpha'\id[\nref] )^{-1}$
	are diagonal and therefore commute with every ${\nref \times \nref}$ matrix
	and the fact that 
	$V (\Lambda^{-1/2} \Lambda^{-1/2}) \ts V^\trans V \ts
	(\Lambda + \alpha'\id[\nref] )^{-1} V^\trans 
	= \nref^{-2} \outgram[Z]^\dagger \ts 
	(\outgram[Z] + \alpha'\id[\nref] )^{-1}$.

	For stability reasons, we can additionally replace $\outgram[Z]^\dagger$ in the above expression
	with its regularized inverse and end up with 
	\begin{equation}
	\restr{(\ecov[Z] + \alpha'\idop[\outrkhs])^{-1}}{\mspan \outreffe} = \nref^{-2} \ts \outreffe \ts 
	(\outgram[Z] + \alpha'\id[\nref] )^{-2} \ts \outreffe^\trans.
	\end{equation}
	
	Here, we make use of the estimate $\eEEO  = \outfe (\ingram[X] + \ndat \alpha \id[\ndat])^{-1} \infe^\trans$
	derived in the literature~\citep{MFSS16} and insert this expression of $\eEEO$ and the above derived expression
	for $\restr{(\ecov[Z] + \alpha'\idop[\outrkhs])^{-1}}{\mspan \outreffe}$ into $\ePEO = (\ecov[Z] + \alpha'\idop[\outrkhs])^{-1} \eEEO$. We discuss
	a potential bias induced by moving from $(\ecov[Z] + \alpha'\idop[\outrkhs])^{-1}$
	to its restriction onto $\mspan \outreffe$ at the end of this subsection. \\
	Inserting both terms yields
	\begin{equation*}
	\ePEO \approx
	\nref^{-2} \ts \outreffe \ts 
	(\outgram[Z] + \alpha'\id[\nref] )^{-2} \ts \outreffe^\trans
	\outfe (\ingram[X] + \ndat \alpha \id[\ndat])^{-1} \infe^\trans,
	\end{equation*}
	which for given $x' \in \inspace$ can be evaluated
	as $\ePEO \ink(x', \cdot) = \sum_{i=1}^{\nref}\beta_i \outk(z_{\scriptscriptstyle i}, \cdot)$
	with the coefficient vector $\beta = \nref^{-2} (\outgram[Z]+ \alpha' \id[\nref])^{-2} \outgram[ZY] 
	(\ingram[X] + \ndat \alpha \id[\ndat])^{-1} [\ink(x_{\scriptscriptstyle 
		1},x'), \dots, \ink(x_{\scriptscriptstyle \ndat}, x')]^\trans \in \R^\nref$.
	The latter is the form presented in the main text.
	
	In general, we introduce a
	bias by replacing $ (\ecov[Z] + \alpha'\idop[\outrkhs])^{-1} $ with 
	its restriction to $\mspan \outreffe$ in 
	the analytical version of the estimate $\ePEO = (\ecov[Z] + \alpha'\idop[\outrkhs])^{-1} \eEEO$.
	This is because
	$\range(\EEO) = \range(\ecov[YX]) = \mspan \outfe $ is
	not necessarily contained in $ \mspan \outreffe$, so information can get ``lost''.  We 
	note that in this general scenario, this cannot be avoided since 
	$ (\ecov[Z] + \alpha'\idop[\outrkhs])^{-1} $ is always of infinite range
	when $\outrkhs$ is infinite dimensional -- however, we must approximate
	$ (\ecov[Z] + \alpha'\idop[\outrkhs])^{-1} $ on 
	the finite-dimensional subspace $\mspan \outreffe$ in numerical scenarios. By assuming that the reference samples are covering the domain $\inspace$ in a sufficient way such that
	this loss of information becomes arbitrarily small,
	replacing $(\ecov[Z] + \alpha'\idop[\outrkhs])^{-1}$ with its restriction
	to $\mspan{\outreffe}$
	also introduces an arbitrarily small error since 
	$(\ecov[Z] + \alpha'\idop[\outrkhs])^{-1}$ is bounded. 
	The detailed analysis of this phenomenon
	will be covered in future work.
	\subsubsection{Closed form expression for mean and variance}
	\label{sec:mean variance closed form}
	Let $\widehat{u} = \sum_{i=1}^{\nref}\beta_i \outk(z_{\scriptscriptstyle i}, \cdot)$ be the RKHS approximation of a density and $\outk(z_{\scriptscriptstyle i}, \cdot)$ be not only a psd kernel evaluated in one argument, but also a probability density with variance $v_\outk$.
	Then the mean of $\widehat{u}$ is given by $m_u = \sum_{i=1}^{\nref}\beta_i z_{\scriptscriptstyle i}$ and the variance by $v_u = \sum_{i=1}^{\nref}\beta_i z^2_{\scriptscriptstyle i} - m_u^2 + v_\outk$.

	\subsection{Computational tricks}
	\label{sec:computational tricks}
	
	In this section, we will detail two tricks that can help fitting large datasets or using density reconstruction when the output domain is high-dimensional.
	
	\subsubsection{Trick for large datasets using factorization of the joint probability}
	\label{sec:Large datasets trick using factorization of the joint probability}
	We fitted the training data of $32\,256$ input-output pairs for the traffic prediction experiment in under $5$ minutes by observing that the dataset only had $1008$ distinct inputs and $32$ output samples per input.
	The following general method takes advantage of this, reducing the involved real matrices from size $32\,256^2$ to $1008^2$.
	Note that the cross-covariance operator can be written as
	$$\cov[\mathit{YX}] = \int_\inspace \psi(y) \otimes \phi(x) \ts \dd \meas_{XY}(x,y) = \int_\inspace  \left ( \int_\outspace \psi(y)  \ts \dd \meas_{Y\mid X=x}(y) \right ) \ts \otimes \phi(x) \ts\dd \meas_{X}(x),$$
	which suggests the empirical estimate $\cov[\mathit{YX}] \approx \ndat^{-1} \sum_{i=1}^\ndat \left ( n_i^{-1} \sum_{j=1}^{n_i} \psi(y_{i,j}) \right ) \otimes \phi(x_i)$,
	where $n_i$ is the number of output samples for input sample $x_i$ and $y_{i,j}$ is the $j$th such sample.
	In feature matrix notation (see \ref{sec:Numerical representation of empirical CDO based on training data}), this is equivalent to $\cov[\mathit{YX}] \approx \ndat^{-1} \outfe\infe^\trans$ for  $\infe = [\ink(x_{\scriptscriptstyle 1},\cdot), \dots, \ink(x_{\scriptscriptstyle \ndat}, \cdot)]$ and
	$\outfe = [n_1^{-1} \sum_{j=1}^{n_1} \outk(y_{\scriptscriptstyle 1, j},\cdot), \dots, n_\ndat^{-1} \sum_{j=1}^{n_\ndat} \outk(y_{\scriptscriptstyle \ndat, j},\cdot)]$.
	For simplicity, consider the conditional mean operator estimate resulting from this.
	This will be given by $\EEO \approx \outfe (\gram[\infe] + \alpha \ndat \id_\ndat)^{-1} \infe^\trans$, where
	$\infe^\trans \infe = \gram[\infe] \in \R^{\ndat \times \ndat}$ is the Gram matrix induced by $\infe$.
	Thus we have to compute the inverse of an $\ndat \times \ndat$ real matrix, while in the standard method a $\left ( \sum_{i=1}^\ndat n_i \right ) \times \left ( \sum_{i=1}^\ndat n_i \right )$ matrix has to be inverted, reducing the complexity from $\mathcal{O}(\ndat^3)$ to $\mathcal{O}\left( \left( \sum_{i=1}^\ndat n_i \right)^3 \right)$.
	When solving the system of equations instead of computing a matrix inverse, we also get computational savings from this trick, even if slightly less so.
	Also, the trick is applicable if there are multiple inputs per output by using the factorizing $\meas_{XY}(x,y) = \meas_{X\mid Y = y}(x) \meas_{Y}(y)$ instead.
	
	\subsubsection{Trick for high dimensions using  Kronecker structure of Gram matrices}
	
	Assume we have a positive definite kernel  $\outk$ over $\R^d$ such that 
	$$\outk([y_1, y_2, \dots, y_d]^\trans, [y'_1, y'_2, \dots, y'_d]^\trans) = \prod_{i=1}^d \outk_i(y_i, y'_i)$$
	
	where $\outk_1, \dots, \outk_d$ are positive definite kernels, i.e., $\outk$ factorizes. 
	Choose $\nref \in \N_+$ such that $\sqrt[d]{\nref}$ is an integer. Furthermore, let $\outgram_i$ be the Gram matrix computed on $\sqrt[d]{\nref}$ samples from the uniform covering the support of the data distribution in dimension $j$. Then $\outgram = \outgram_1 \otimes \dots \otimes \outgram_d$ and by properties of the Kronecker product, we have $\outgram^\inv = \outgram_1^\inv \otimes \dots \otimes \outgram_d^\inv$.\\
	Thus, by inverting $d$ gram matrices of size $\sqrt[d]{\nref} \times \sqrt[d]{\nref}$ and computing Kronecker products, we can get the inverse of an $\nref \times \nref$ gram matrix. The inversion has computational complexity $\mathcal{O}(d{\nref}^{3/d})$, while the Kronecker products have complexity $ \mathcal{O}\left(\left(\sqrt[d]{\nref}\right) ^{2d}\right) = \mathcal{O}(\nref^2)$.
	Assuming $d \geq 2$ and $\sqrt[d]{\nref} > 2$, the $ \mathcal{O}(\nref^2) $ complexity of the Kronecker products will dominate. This is a significant improvement from the $ \mathcal{O}(\nref^3) $ computational complexity it would take to invert $\outgram$ directly.
	The $d$-dimensional  points for which $\outgram$ is the Gram matrix uniformly cover a $d$-dimensional box.
	Thus, this trick will be useful with a Lebesgue (i.e., uniform) reference measure on this box.
	Another advantage is that the computation of Kronecker products is vectorized in most linear algebra packages and trivial to parallelize across dimensions, and further computation could be saved by taking advantage of the symmetry of Gram matrices when computing Kronecker products.
	Similar tricks have been used in the literature on scalable Gaussian Processes, see for example \cite{WN15,F15,NGLOR15,EN18}.

\end{document}